\documentclass[a4paper]{article}
\usepackage{INTERSPEECH2021}
\usepackage{hyperref}
\usepackage{url}
\usepackage{booktabs}       
\usepackage{amsfonts}       
\usepackage[dvipsnames]{xcolor} 

\makeatletter
\def\blfootnote{\xdef\@thefnmark{}\@footnotetext}
\makeatother

\title{FLEURS: Few-shot Learning Evaluation of\\
Universal Representations of Speech}
\name{Author Name$^1$, Co-author Name$^2$}
\name{
Alexis Conneau\textsuperscript{\dag\ddag 1}\email{aconneau@google.com},
Min Ma\textsuperscript{\dag 2},
Simran Khanuja\textsuperscript{\dag 2},
Yu Zhang\textsuperscript{2}\email{ngyuzh@google.com},
Vera Axelrod\textsuperscript{2},
Siddharth Dalmia\textsuperscript{3},
Jason~Riesa\textsuperscript{2}\email{riesa@google.com},
Clara Rivera\textsuperscript{2},
Ankur Bapna\textsuperscript{\ddag 2}\email{ankurbpn@google.com}
\email{aconneau@fb.com, \{minm,simrankh,ngyuzh,vaxelrod,ankurbpn\}@google.com, sdalmia@cs.cmu.edu}
}
\address{\textsuperscript{1}Meta AI Research, \textsuperscript{2}Google Research, \textsuperscript{3}Carnegie Mellon University}

\newcommand{\wvbert}{w2v-bert-51 (0.6B)}

\newcommand{\mctc}{mSLAM}

\newcommand{\mctcb}[1]{\mctc{} {(#1B)}}

\newcommand{\insertcorpora}{
\begin{table*}[hbt!]
\scriptsize
\centering
\caption{Compare FLEURS to common public multilingual speech benchmarks.}
\label{tab:corpora}

\begin{tabular}{l|cccccccc}
\toprule
Dataset &
\#Languages &
Total Duration &
Domains &
Speech Type &
Transcripts &
Parallel text &
Parallel speech &\\
\midrule
BABEL~\cite{gales2014babel} &
17 &
1k hours &
Conversational &
Spontaneous &
Yes &
No &
No & \\
CommonVoice~\cite{ardila2019common} &
93 &
15k hours &
Open domain &
Read &
Yes &
No &
No\\
CMU Wilderness~\cite{black2019cmu} &
700 &
14k hours &
Religion &
Read &
Yes &
Yes &
Yes\\
MLS~\cite{pratap2020mls} & 
8 &
50.5k hours &
Audiobook &
Read &
Yes &
No &
No & \\
CoVoST-2~\cite{wang2020covost2} &
22 &
2.9k hours &
Open domain &
Read &
Yes &
Yes &
No & \\
Voxlingua-107~\cite{valk2020voxlingua107} &
107 &
6.6k hours &
YouTube &
Spontaneous &
No &
No &
No\\
Europarl-ST~\cite{iranzo2020europarl} & 
6 &
500 hours &
Parliament &
Spontaneous &
Yes &
Yes &
No & \\
MuST-C~\cite{cattoni2021must} &
9 &
385 hours &
TED talks &
Spontaneous &
Yes &
Yes &
No & \\
mTEDx~\cite{salesky2021multilingual} & 
9 &
1k hours &
TED talks &
Spontaneous &
Yes &
Yes &
No & \\
VoxPopuli~\cite{wang2021voxpopuli} &
24 &
400k hours &
Parliament &
Spontaneous &
Partial &
Partial &
Partial & \\
CVSS~\cite{jia2022cvss} & 
22 &
1.1k hours &
Open domain &
Read/Synthetic &
Yes &
Yes &
Yes & \\
FLEURS (this work) & 
102 &
1.4k hours &
Wikipedia &
Read &
Yes &
Yes &
Yes\\
\bottomrule
\end{tabular}  
\caption{A comparison of commonly used datasets for multilingual speech representation learning, ASR, Speech Translation and Speech-LangID. CommonVoice statistics as on 24th May 2022.}
\end{table*}
}

\newcommand{\insertfleursbasics}{
 \begin{table}[h!]
        \begin{center}
            \captionof{table}{Statistics for speech and transcript data in FLEURS.
            \label{tab:fleursbasics}}
            \resizebox{\linewidth}{!}{
            \begin{tabular}[t]{l|ccccccc|c}
            \toprule
            {  Data Statistics} & WE & EE & CMN & SSA & SA & SEA & CJK & All \\
                \midrule
                \midrule
                train speech hours & 231h & 134h & 116h & 237h & 124h & 112h & 32h & 987h \\
                dev speech hours & 29h & 18h & 14h & 24h & 16h & 14h & 4h & 120h \\
                test speech hours & 68h & 43h & 33h & 58h & 37h & 35h & 9h & 283h\\
                train transcript tokens & 1475k & 772k & 630k & 1072k & 699k & 525k & 405k & 5578k\\
                dev transcript tokens & 184k & 107k & 75k & 116k & 93k & 65k & 51k & 692k \\
                test transcript tokens & 443k & 260k & 181k & 272k & 210k & 158k & 116k & 1640k\\
                \bottomrule
            \end{tabular}
            }
        \end{center}
    \end{table}
}

\newcommand{\insertFLEURS}{
\begin{table}[h!]
\begin{tabular}{l|rrrrrr}
\toprule

ISO 639-3 & Language & Family & Subgrouping & Script & Bitext & Mono \\
afr & Afrikaans & Indo-European & Germanic & Latin & 570K & 26.1M \\
amh & Amharic & Afro-Asiatic & Afro-Asiatic & Ge’ez & 339K & 3.02M \\
ara & Arabic & Afro-Asiatic & Afro-Asiatic & Arabic & 25.2M & 126M \\
hye & Armenian & Indo-European & Other IE & Armenian & 977K & 25.4M \\
asm & Assamese & Indo-European & Indo-Aryan & Bengali & 43.7K & 738K \\
ast & Asturian & Indo-European & Romance & Latin & 124K & — \\
azj & Azerbaijani & Turkic & Turkic & Latin & 867K & 41.4M \\
bel & Belarusian & Indo-European & Balto-Slavic & Cyrillic & 42.4K & 24M \\
ben & Bengali & Indo-European & Indo-Aryan & Bengali & 2.16M & 57.9M \\
bos & Bosnian & Indo-European & Balto-Slavic & Latin & 187K & 15.9M \\
bul & Bulgarian & Indo-European & Balto-Slavic & Cyrillic & 10.3M & 235M \\
mya & Burmese & Sino-Tibetan & Sino-Tibetan+Kra-Dai & Myanmar & 283K & 2.66M \\
cat & Catalan & Indo-European & Romance & Latin & 5.77M & 77.7M \\
ceb & Cebuano & Austronesian & Austronesian & Latin & 484K & 4.11M \\
cmn & Mandarin Chinese & Sino-Tibetan & Sino-Tibetan+Kra-Dai & Han & 37.9M & 209M \\
yue & Cantonese Chinese & Sino-Tibetan & Sino-Tibetan+Kra-Dai & Han & 37.9M & 85.2M \\
hrv & Croatian & Indo-European & Balto-Slavic & Latin & 42.2K & 144M \\
ces & Czech & Indo-European & Balto-Slavic & Latin & 23.2M & 124M \\
dan & Danish & Indo-European & Germanic & Latin & 10.6M & 344M \\
nld & Dutch & Indo-European & Germanic & Latin & 82.4M & 230M \\
est & Estonian & Uralic & Uralic & Latin & 4.82M & 46M \\
tgl & Filipino (Tagalog) & Austronesian & Austronesian & Latin & 70.6K & 107M \\
fin & Finnish & Uralic & Uralic & Latin & 15.2M & 377M \\
fra & French & Indo-European & Romance & Latin & 289M & 428M \\
ful & Fula & Atlantic-Congo & Nilotic+Other AC & Latin & 71K & 531K \\
glg & Galician & Indo-European & Romance & Latin & 1.13M & 4.22M \\
lug & Ganda & Atlantic-Congo & Bantu & Latin & 14.4K & 537K \\
kat & Georgian & Kartvelian & Other & Georgian & 1.23M & 31.7M \\
deu & German & Indo-European & Germanic & Latin & 216M & 417M \\
ell & Greek & Indo-European & Other IE & Greek & 23.7M & 201M \\
guj & Gujarati & Indo-European & Indo-Aryan & Gujarati & 160K & 9.41M \\
hau & Hausa & Afro-Asiatic & Afro-Asiatic & Latin & 335K & 5.87M \\
heb & Hebrew & Afro-Asiatic & Afro-Asiatic & Hebrew & 6.64M & 208M \\
hin & Hindi & Indo-European & Indo-Aryan & Devanagari & 3.3M & 104M \\
hun & Hungarian & Uralic & Uralic & Latin & 16.3M & 385M \\
isl & Icelandic & Indo-European & Germanic & Latin & 1.17M & 37.5M \\
ibo & Igbo & Atlantic-Congo & Nilotic+Other AC & Latin & 145K & 693K \\
ind & Indonesian & Austronesian & Austronesian & Latin & 39.1M & 1.05B \\
gle & Irish & Indo-European & Other IE & Latin & 329K & 1.54M \\
ita & Italian & Indo-European & Romance & Latin & 116M & 179M \\
jpn & Japanese & Japonic & Other & Han, Hirgana, Katakana & 23.2M & 458M \\
jav & Javanese & Austronesian & Austronesian & Latin & 1.49M & 24.4M \\
kea & Kabuverdianu & Indo-European & Romance & Latin & 5.46K & 178K \\
kam & Kamba & Atlantic-Congo & Bantu & Latin & 50K & 181K \\
kan & Kannada & Dravidian & Dravidian & Telugu-Kannada & 155K & 13.1M \\
kaz & Kazakh & Turkic & Turkic & Cyrillic & 701K & 35.6M \\
khm & Khmer & Austro-Asiatic & Austro-Asiatic & Khmer & 398K & 8.87M \\
kor & Korean & Koreanic & Other & Hangul & 7.46M & 390M \\
kir & Kyrgyz & Turkic & Turkic & Cyrillic & 566K & 2.02M \\
lao & Lao & Kra-Dai & Sino-Tibetan+Kra-Dai & Lao & 153K & 2.47M \\
lav & Latvian & Indo-European & Balto-Slavic & Latin & 4.8M & 68.4M \\
lin & Lingala & Atlantic-Congo & Bantu & Latin & 21.1K & 336K \\
lit & Lithuanian & Indo-European & Balto-Slavic & Latin & 6.69M & 111M \\
luo & Luo & Nilo-Saharan & Nilotic+Other AC & Latin & 142K & 239K \\
ltz & Luxembourgish & Indo-European & Germanic & Latin & 3.41M & — \\
mkd & Macedonian & Indo-European & Balto-Slavic & Cyrillic & 1.13M & 28.8M \\
msa & Malay & Austronesian & Austronesian & Latin & 968K & 77.5M \\
mal & Malayalam & Dravidian & Dravidian & Malayalam & 497K & 24.8M \\
mlt & Maltese & Afro-Asiatic & Afro-Asiatic & Latin & 5.82M & — \\
mri & Maori & Austronesian & Austronesian & Latin & 196K & — \\
mar & Marathi & Indo-European & Indo-Aryan & Devanagari & 109K & 14.4M \\
mon & Mongolian & Mongolic & Other & Cyrillic & 555K & 20.4M \\
npi & Nepali & Indo-European & Indo-Aryan & Devanagari & 19.6K & 17.9M \\
nso & Northern Sotho & Atlantic-Congo & Bantu & Latin & 13.8K & 612K \\
nob & Norwegian & Indo-European & Germanic & Latin & 10.9M & 338M \\
nya & Nyanja & Atlantic-Congo & Bantu & Latin & 932K & — \\
oci & Occitan & Indo-European & Romance & Latin & 5.11K & — \\
ory & Oriya & Indo-European & Indo-Aryan & Oriya & 5K & 2.47M \\
orm & Oromo & Afro-Asiatic & Afro-Asiatic & Latin & 162K & 752K \\
pus & Pashto & Indo-European & Indo-Aryan & Perso-Arabic & 293K & 12M \\
fas & Persian & Indo-European & Indo-Aryan & Perso-Arabic & 6.63M & 611M \\
pol & Polish & Indo-European & Balto-Slavic & Latin & 40.9M & 256M \\
por & Portuguese (Brazil) & Indo-European & Romance & Latin & 137M & 340M \\
pan & Punjabi & Indo-European & Indo-Aryan & Gurmukhi & 142K & 5.02M \\
ron & Romanian & Indo-European & Romance & Latin & 31.9M & 391M \\
rus & Russian & Indo-European & Balto-Slavic & Cyrillic & 127M & 849M \\
srp & Serbian & Indo-European & Balto-Slavic & Cyrillic & 7.01M & 35.7M \\
sna & Shona & Atlantic-Congo & Bantu & Latin & 877K & — \\
snd & Sindhi & Indo-European & Indo-Aryan & Perso-Arabic & 21.8K & 314K \\
slk & Slovak & Indo-European & Balto-Slavic & Latin & 10.5M & 174M \\
slv & Slovenian & Indo-European & Balto-Slavic & Latin & 5.42M & 74.7M \\
som & Somali & Afro-Asiatic & Afro-Asiatic & Latin & 358K & 14.1M \\
ckb & Sorani Kurdish & Indo-European & Indo-Aryan & Arabic & 305K & 7.98M \\
spa & Spanish (Latin America) & Indo-European & Romance & Latin & 315M & 379M \\
swh & Swahili & Atlantic-Congo & Bantu & Latin & 349K & 35.8M \\
swe & Swedish & Indo-European & Germanic & Latin & 54.8M & 580M \\
tgk & Tajik & Indo-European & Indo-Aryan & Cyrillic & 544K & — \\
tam & Tamil & Dravidian & Dravidian & Tamil & 992K & 68.2M \\
tel & Telugu & Dravidian & Dravidian & Telugu-Kannada & 381K & 17.2M \\
tha & Thai & Kra-Dai & Sino-Tibetan+Kra-Dai & Thai & 10.6M & 319M \\
tur & Turkish & Turkic & Turkic & Latin & 41.2M & 128M \\
ukr & Ukrainian & Indo-European & Balto-Slavic & Cyrillic & 5.44M & 357M \\
umb & Umbundu & Atlantic-Congo & Bantu & Latin & 217K & 142K \\
urd & Urdu & Indo-European & Indo-Aryan & Perso-Arabic & 630K & 28M \\
uzb & Uzbek & Turkic & Turkic & Latin & — & 7.54M \\
vie & Vietnamese & Austro-Asiatic & Austro-Asiatic & Latin & 32.1M & 992M \\
cym & Welsh & Indo-European & Other IE & Latin & 826K & 12.7M \\
wol & Wolof & Atlantic-Congo & Nilotic+Other AC & Latin & 86.9K & 676K \\
xho & Xhosa & Atlantic-Congo & Bantu & Latin & 130K & 995K \\
yor & Yoruba & Atlantic-Congo & Nilotic+Other AC & Latin & 171K & 1.59M \\
zul & Zulu & Atlantic-Congo & Bantu & Latin & 123K & 994K \\
\caption{101 languages in FLEURS statistics and table of results}
\label{tab:covost_enx_full}
\end{table}
}

\newcommand{\insertfleursfullasr}{
\begin{table*}[hbt!]
\scriptsize
\centering
\caption{FLEURS full ASR results, reporting \% CER ($\downarrow$), for all 102 languages.}
\label{tab:fleurs_full}

\begin{tabular}{l|rrrrrrrrrrrrrrr}
\toprule
& \multicolumn{15}{c}{Western European}\\
\midrule
Language & ast & bs &  ca &  hr &  da &  nl &  en &  fi &  fr &  gl &  de &  el &  hu &  is & ga \\
\midrule 
\wvbert & 8.7 & 5.8 & 4.3 & 9.3 & 11.3 & 6.0 & 17.2 & 3.0 & 9.6 & 8.6 & 8.0 & 11.7 & 24.9 & 11.9 & 39.5 \\
\mctcb{0.6} & 7.5 & 5.1 & 4.7 & 8.5 & 14.0 & 6.8 & 16.3 & 3.4 & 9.7 & 8.7 & 5.7 & 12.0 & 18.1 & 12.8 & 40.5 \\
\end{tabular}  

\begin{tabular}{l|rrrrrrrrrrrrrrr}
\toprule
& \multicolumn{10}{c|}{Western European (WE)} & \multicolumn{5}{c}{Eastern European} \\
\midrule
Language & it &  kea &  lb &  mt &  nb &  oc &  pt &  es &  sv & cy & hy & be & bg & cs & et \\
\midrule 
\wvbert & 2.6 & 4.9 & 19.4 & 17.3 & 5.8 & 11.7 & 4.2 & 3.7 & 7.6 & 11.1 & 7.2 & 9.1 & 4.8 & 10.3 & 3.1 \\
\mctcb{0.6} & 2.3 & 5.1 & 21.0 & 17.3 & 6.1 & 12.7 & 4.4 & 3.3 & 7.8 & 12.0 & 7.8 & 7.5 & 5.2 & 9.2 & 3.5 \\
\end{tabular}  

\begin{tabular}{l|rrrrrrrrrrrrrrr}
\toprule
& \multicolumn{11}{c|}{Eastern European (EE)} & \multicolumn{4}{c}{Central-Asia and}\\
\midrule
Language & ka & lv & lt & mk & pl & ro & ru & sr & sk & sl & uk & ar & az & he & kk \\
\midrule 
\wvbert & 30.7 & 4.4 & 12.8 & 11.8 & 5.0 & 8.0 & 5.6 & 11.6 & 4.9 & 7.9 & 21.4 & 10.5 & 12.7 & 37.2 & 6.5 \\
\mctcb{0.6} & 31.0 & 4.5 & 11.6 & 9.8 & 6.3 & 8.4 & 6.6 & 12.2 & 4.8 & 10.3 & 21.4 & 11.0 & 15.9 & 42.5 & 5.7 \\
\end{tabular}  

\begin{tabular}{l|rrrrrrrrrrrrrrr}
\toprule
& \multicolumn{8}{c|}{Middle-East and North-Africa (CMN)} & \multicolumn{7}{c}{Sub-Saharan Africa}\\
\midrule
Language & ky & mn & ps & fa & ckb & tg & tr & uz & af & am & ff & lg & ha & ig & kam \\
\midrule 
\wvbert & 8.3 & 15.2 & 20.4 & 15.7 & 15.1 & 7.1 & 8.5 & 16.8 & 9.5 & 17.2 & 27.8 & 12.4 & 9.8 & 18.1 & 13.5 \\
\mctcb{0.6} & 8.0 & 16.1 & 21.1 & 10.0 & 15.0 & 7.6 & 9.7 & 15.5 & 11.9 & 17.8 & 27.5 & 12.9 & 10.5 & 18.7 & 14.0 \\
\end{tabular}  

\begin{tabular}{l|rrrrrrrrrrrrrrr}
\toprule
& \multicolumn{13}{c|}{Sub-Saharan Africa (SSA)} & \multicolumn{2}{c}{South-Asia} \\
\midrule
Language & ln & luo & nso & ny & om & sn & so & sw & umb & wo & xh & yo & zu & as & bn \\
\midrule 
\wvbert & 6.1 & 7.0 & 11.7 & 11.5 & 21.7 & 16.6 & 21.3 & 19.4 & 13.1 & 17.8 & 23.9 & 23.3 & 9.8 & 13.7 & 9.4 \\
\mctcb{0.6} & 6.8 & 7.4 & 11.9 & 12.4 & 22.6 & 17.6 & 23.4 & 20.2 & 14.0 & 18.8 & 25.1 & 23.2 & 10.8 & 14.0 & 9.7 \\
\end{tabular}  

\begin{tabular}{l|rrrrrrrrrrrrrrr}
\toprule
& \multicolumn{12}{c|}{South-Asia (SA)} & \multicolumn{3}{c}{South-East Asia}\\
\midrule
Language & gu & hi & kn & ml & mr & ne & or & pa & sd & ta & te & ur & my & ceb & fil\\
\midrule 
\wvbert & 9.3 & 12.4 & 7.0 & 8.6 & 14.8 & 13.0 & 19.2 & 13.6 & 16.0 & 11.8 & 12.0 & 82.9 & 18.2 & 5.9 & 5.6 \\
\mctcb{0.6} & 9.6 & 15.2 & 9.6 & 12.2 & 18.9 & 14.8 & 20.7 & 15.2 & 20.8 & 13.2 & 12.3 & 83.1 & 18.8 & 6.2 & 5.5 \\
\end{tabular}  

\begin{tabular}{l|rrrrrrrrrrrrrrr}
\toprule
& \multicolumn{8}{c|}{South-East Asia (SEA)} & \multicolumn{4}{c|}{CJK} & & & \\
\midrule
Language & id & jv & km & lo & ms & mi & th & vi & yue & cmn & ja & ko & All & & \\
\midrule 
\wvbert & 5.2 & 7.0 & 29.9 & 38.1 & 8.6 & 10.3 & 18.6 & 14.2 & 37.0 & 22.2 & 37.7 & 21.7 & 14.1 \\
\mctcb{0.6} & 5.6 & 7.1 & 30.2 & 37.5 & 7.2 & 11.2 & 20.1 & 14.3 & 39.8 & 23.1 & 39.2 & 22.4
 & 14.6 &  \\
\bottomrule
\end{tabular}

\end{table*}
}

\newcommand{\insertsc}{
 \begin{table}[bht!]
        \begin{center}
            \captionof{table}{\textbf{  Speech identification }- FLEURS langID baselines, reporting \% accuracy ($\uparrow$), by geographical group.
            \label{tab:fleurssid}}
            \resizebox{\linewidth}{!}{
            \begin{tabular}[t]{l|ccccccc|c}
            \toprule
            {  Model} & WE & EE & CMN & SSA & SA & SEA & CJK & Avg. \\
                \midrule
                \midrule
                \multicolumn{9}{l}{\it Speech identification accuracy for all languages} \\
                \midrule
                \multicolumn{1}{l|}{\# languages} & 25 & 16 & 12 & 20 & 14 & 11 & 4 & 102 \\
                \midrule
                \wvbert  & 85.3 & 78.4 & 72.9 & 59.1 & 52.0 & 65.7 & 89.7 & 71.4 \\
                
                \mctcb{0.6} & 84.6 & 81.3 & 75.9 & 62.2 & 51.7 & 73.4 & 87.8 & 73.3 \\
                \bottomrule
            \end{tabular}
            }
        \end{center}
    \end{table}
}

\newcommand{\insertfleursasrseenvsunseen}{
 \begin{table}[bht!]
        \begin{center}
            \captionof{table}{\textbf{  Speech recognition }- Fleurs massively multilingual ASR baselines, reporting \% CER ($\downarrow$), by geographical group.
            \label{tab:fleurs}}
            \resizebox{\linewidth}{!}{
            \begin{tabular}[t]{l|ccccccc|c}
            \toprule
            {  Model} & WE & EE & CMN & SSA & SA & SEA & CJK & Avg. \\
                \midrule
                \midrule
                \multicolumn{9}{l}{\it Speech recognition CER for all languages} \\
                \midrule
                \multicolumn{1}{l|}{\# languages} & 25 & 16 & 12 & 20 & 14 & 11 & 4 & 102 \\
                \midrule
                \wvbert  & 10.7 & 9.9 & 14.5 & 15.6 & 17.4 & 14.7 & 24.6 & 14.1 \\
                \mctcb{0.6} & 10.6 & 10.0 & 14.8 & 16.4 & 19.2 & 14.9 & 25.0 & 14.6 \\
                \bottomrule
            \end{tabular}
            }
        \end{center}
    \end{table}
}

\newcommand{\insertlanguages}{
\begin{table*}[ht!]
\begin{center}
\captionof{table}{\textbf{Characteristics of the 102 languages in FLEURS}, with their ISO codes, language families, estimated number of speakers in millions (\#S) from~\cite{Esch2022}.
Languages are grouped geographically in Western Europe (WE), Eastern Europe (EE), Central-Asia/Middle-East/North-Africa (CMN), Sub-Saharan Africa (SSA), South Asia (SA), South-East Asia (SEA) and CJK languages. Note that we make some minor changes compared to FLoRes, renaming ``Chinese (Simp)'' to ``Mandarin Chinese'' with code ``cmn'' (not ``zho'') and renaming ``Chinese (Trad)'' to ``Cantonese Chinese" with code ``yue'' (not ``zho''). 
\label{tab:langs}}
\begin{tabular}[t]{|r|l|l|l|l|l|r|}
\toprule
Idx & Language & ISO 639-3 & ISO 639-1 & Family & Group & \#S \\ 
\midrule
1 & Afrikaans & afr & af & Indo-European & SSA & 17  \\ 
2 & Amharic & amh & am & Afro-Asiatic & SSA & 22  \\ 
3 & Arabic & ara & ar & Afro-Asiatic & CMN & 180  \\ 
4 & Armenian & hye & hy & Indo-European & EE & 6  \\ 
5 & Assamese & asm & as& Indo-European & SA & 13  \\ 
6 & Asturian & ast & - & Indo-European & WE & 0.6  \\ 
7 & Azerbaijani & azj & az & Turkic & CMN & 18  \\ 
8 & Belarusian & bel & be & Indo-European & EE & 3  \\ 
9 & Bengali & ben & bn & Indo-European & SA & 260  \\ 
10 & Bosnian & bos & bs& Indo-European & WE & 9  \\ 
11 & Bulgarian & bul & bg & Indo-European & EE & 7 \\ 
12 & Burmese & mya & my & Sino-Tibetan & SEA & 33  \\ 
13 & Cantonese Chinese & yue & - & Sino-Tibetan & CJK & 920  \\ 
14 & Catalan & cat & ca & Indo-European & WE & 4  \\ 
15 & Cebuano & ceb & - & Austronesian & SEA & 16  \\ 
16 & Croatian & hrv & hr & Indo-European & WE & 4  \\ 
17 & Czech & ces & cs & Indo-European & EE & 10  \\ 
18 & Danish & dan & da & Indo-European & WE & 5  \\ 
19 & Dutch & nld & nl & Indo-European & WE & 21  \\ 
20 & English & eng & en & Indo-European & WE & 550   \\
21 & Estonian & est & et & Uralic & EE & 1  \\ 
22 & Filipino (Tagalog) & tgl & tg & Austronesian & SEA & 22  \\ 
23 & Finnish & fin & fi & Uralic & WE & 5  \\ 
24 & French & fra & fr & Indo-European & WE & 280  \\ 
25 & Fula & ful & ff & Atlantic-Congo & SSA & 12  \\ 
26 & Galician & glg & gl & Indo-European & WE & 2  \\ 
27 & Ganda & lug & lg & Atlantic-Congo & SSA & 4  \\ 
28 & Georgian & kat & ka & Kartvelian & EE & 4  \\ 
29 & German & deu & de & Indo-European & WE & 83  \\ 
30 & Greek & ell & el &  Indo-European & WE & 13  \\ 
31 & Gujarati & guj & gu & Indo-European & SA & 56  \\ 
32 & Hausa & hau & ha &  Afro-Asiatic & SSA & 70  \\ 
33 & Hebrew & heb & he & Afro-Asiatic & CMN & 4  \\ 
34 & Hindi & hin & hi & Indo-European & SA & 320  \\ 
35 & Hungarian & hun & hu & Uralic & WE & 13  \\ 
36 & Icelandic & isl & is & Indo-European & WE &  0.3  \\ 
37 & Igbo & ibo & ig & Atlantic-Congo & SSA & 18  \\ 
38 & Indonesian & ind & id & Austronesian & SEA & 200  \\ 
39 & Irish & gle & ga & Indo-European & WE & 0.2  \\ 
40 & Italian & ita & it & Indo-European & WE & 61  \\ 
41 & Japanese & jpn & ja & Japonic & CJK & 130  \\ 
42 & Javanese & jav & jv & Austronesian & SEA & 85  \\ 
43 & Kabuverdianu & kea & - & Indo-European & WE & 0.9  \\ 
44 & Kamba & kam & - & Atlantic-Congo & SSA & 4  \\ 
45 & Kannada & kan & kn & Dravidian & SA & 43  \\ 
46 & Kazakh & kaz & kk & Turkic & CMN & 11  \\ 
47 & Khmer & khm & km & Austro-Asiatic & SEA & 16  \\ 
48 & Korean & kor & ko & Koreanic & CJK & 52  \\ 
49 & Kyrgyz & kir & ky & Turkic & CMN & 8  \\ 
50 & Lao & lao & lo & Kra-Dai & SEA & 20  \\ 
51 & Latvian & lav & lv & Indo-European & EE & 2  \\ 
\bottomrule
\end{tabular}
\end{center}
\end{table*}

\begin{table*}[ht!]
\begin{center}
\begin{tabular}[t]{|r|l|l|l|l|l|r|}
\toprule
Idx & Language & ISO 639-3 & ISO 639-1 & Family & Group & \#S \\ 
\midrule
52 & Lingala & lin & ln & Atlantic-Congo & SSA & 15   \\
53 & Lithuanian & lit & lt & Indo-European & EE & 2   \\
54 & Luo & luo & - & Nilo-Saharan & SSA & 4   \\
55 & Luxembourgish & ltz & lb & Indo-European & WE & 0.4   \\
56 & Macedonian & mkd & mk & Indo-European & EE & 1   \\
57 & Malay & msa & ms & Austronesian & SEA & 80   \\
58 & Malayalam & mal & ml & Dravidian & SA & 77   \\
59 & Maltese & mlt & mt & Afro-Asiatic & WE & 0.5   \\
60 & Mandarin Chinese & cmn & - & Sino-Tibetan & CJK & 80   \\
61 & Maori & mri & mi & Austronesian & SEA & 0.2   \\
62 & Marathi & mar & mr & Indo-European & SA & 83   \\
63 & Mongolian & mon & mn & Mongolic & CMN & 5   \\
64 & Nepali & npi & ne & Indo-European & SA & 16   \\
65 & Northern Sotho & nso & - & Atlantic-Congo & SSA & 14   \\
66 & Norwegian & nob & nb & Indo-European & WE & 5   \\
67 & Nyanja & nya & ny & Atlantic-Congo & SSA & 12   \\
68 & Occitan & oci & oc & Indo-European & WE & 0.5   \\
69 & Oriya & ory & or & Indo-European & SA & 35   \\
70 & Oromo & orm & om & Afro-Asiatic & SSA & 24   \\
71 & Pashto & pus & ps & Indo-European & CMN & 13   \\
72 & Persian & fas & fa & Indo-European & CMN & 40   \\
73 & Polish & pol & pl & Indo-European & EE & 38   \\
74 & Portuguese (Brazil) & por & pt & Indo-European & WE & 220   \\
75 & Punjabi & pan & pa & Indo-European & SA & 113   \\
76 & Romanian & ron & ro & Indo-European & EE & 19   \\
77 & Russian & rus & ru & Indo-European & EE & 150   \\
78 & Serbian & srp & sr & Indo-European & EE & 6   \\
79 & Shona & sna & sn & Atlantic-Congo & SSA & 9   \\
80 & Sindhi & snd & sd & Indo-European & SA & 68   \\
81 & Slovak & slk & sk & Indo-European & EE & 4   \\
82 & Slovenian & slv & sl & Indo-European & EE & 2   \\
83 & Somali & som & so & Afro-Asiatic & SSA & 24   \\
84 & Sorani Kurdish & ckb & - & Indo-European & CMN &  7   \\
85 & Spanish & spa & es & Indo-European & WE & 490   \\
86 & Swahili & swh & sw & Atlantic-Congo & SSA & 24   \\
87 & Swedish & swe & sv & Indo-European & WE & 8   \\
88 & Tajik & tgk & tg & Indo-European & CMN & 8   \\
89 & Tamil & tam & ta & Dravidian & SA & 76   \\
90 & Telugu & tel & te & Dravidian & SA & 82   \\
91 & Thai & tha & th & Kra-Dai & SEA & 20   \\
92 & Turkish & tur & tr & Turkic & CMN & 82   \\
93 & Ukrainian & ukr & uk & Indo-European & EE & 32   \\
94 & Umbundu & umb & - & Atlantic-Congo & SSA & 6   \\
95 & Urdu & urd & ur & Indo-European & SA & 120   \\
96 & Uzbek & uzb & uz & Turkic & CMN & 57   \\
97 & Vietnamese & vie & vi & Austro-Asiatic & SEA & 96   \\
98 & Welsh & cym & cy & Indo-European & WE & 0.7   \\
99 & Wolof & wol & wo & Atlantic-Congo & SSA & 4   \\
100 & Xhosa & xho & xh & Atlantic-Congo & SSA & 19   \\
101 & Yoruba & yor & yo & Atlantic-Congo & SSA & 21   \\
102 & Zulu & zul & zu & Atlantic-Congo & SSA & 11  \\ 
\bottomrule
\end{tabular}
\end{center}
\end{table*}

}

\newcommand{\insertfleursspeechtextretrieval}{
\begin{table*}[hbt!]
\scriptsize
\centering
\caption{FLEURS Speech-to-Text and Text-to-Speech retrieval results, reporting \% P@1 ($\uparrow$), for all 102 languages. }
\label{tab:fleurs_full_retrieval}

\begin{tabular}{l|rrrrrrrrrrrrrrr}
\toprule
& \multicolumn{15}{c}{Western European}\\
\midrule
Language & ast & bs &  ca &  hr &  da &  nl &  en &  fi &  fr &  gl &  de &  el &  hu &  is & ga \\
\midrule 
Speech-to-Text Retrieval & 90.1 & 95.5 & 93.2 & 98.0 & 94.1 & 95.3 & 96.0 & 93.0 & 90.7 & 90.9 & 91.2 & 81.2 & 85.3 & 71.7 & 55.1 \\
Text-to-Speech Retrieval & 83.5 & 95.1 & 87.0 & 95.3 & 90.1 & 87.1 & 94.4 & 90.6 & 80.6 & 81.0 & 81.9 & 74.6 & 88.4 & 67.4 & 56.7 \\

\end{tabular}  

\begin{tabular}{l|rrrrrrrrrrrrrrr}
\toprule
& \multicolumn{10}{c|}{Western European (WE)} & \multicolumn{5}{c}{Eastern European} \\
\midrule
Language & it &  kea &  lb &  mt &  nb &  oc &  pt &  es &  sv & cy & hy & be & bg & cs & et \\
\midrule 
Speech-to-Text Retrieval & 93.5 & 95.4 & 80.5 & 92.7 & 91.9 & 77.4 & 91.9 & 69.6 & 94.2 & 82.3 & 50.3 & 90.2 & 95.1 & 98.1 & 95.6 \\
Text-to-Speech Retrieval & 86.0 & 95.6 & 78.8 & 89.0 & 85.7 & 85.2 & 88.9 & 61.5 & 90.4 & 79.6 & 52.5 & 85.5 & 92.1 & 96.3 & 94.1 \\
\end{tabular}  

\begin{tabular}{l|rrrrrrrrrrrrrrr}
\toprule
& \multicolumn{11}{c|}{Eastern European (EE)} & \multicolumn{4}{c}{Central-Asia and}\\
\midrule
Language & ka & lv & lt & mk & pl & ro & ru & sr & sk & sl & uk & ar & az & he & kk \\
\midrule 
Speech-to-Text Retrieval & 70.5 & 97.4 & 96.8 & 96.1 & 95.8 & 92.0 & 93.2 & 97.7 & 97.6 & 97.4 & 93.5 & 82.7 & 83.0 & 64.0 & 88.7 \\
Text-to-Speech Retrieval & 66.9 & 95.7 & 94.0 & 92.7 & 92.0 & 86.4 & 89.2 & 95.6 & 95.6 & 94.8 & 88.9 & 82.7 & 75.2 & 71.3 & 89.7 \\

\end{tabular}  

\begin{tabular}{l|rrrrrrrrrrrrrrr}
\toprule
& \multicolumn{8}{c|}{Middle-East and North-Africa (CMN)} & \multicolumn{7}{c}{Sub-Saharan Africa}\\
\midrule
Language & ky & mn & ps & fa & ckb & tg & tr & uz & af & am & ff & lg & ha & ig & kam \\
\midrule 
Speech-to-Text Retrieval & 84.3 & 70.7 & 84.8 & 85.4 & 80.8 & 76.3 & 84.5 & 67.6 & 90.1 & 34.1 & 81.4 & 90.7 & 84.5 & 85.8 & 89.7 \\
Text-to-Speech Retrieval & 78.6 & 70.4 & 78.7 & 85.5 & 76.9 & 70.7 & 81.0 & 63.5 & 86.7 & 40.3 & 65.8 & 90.6 & 85.8 & 90.2 & 90.7 \\
\end{tabular}  

\begin{tabular}{l|rrrrrrrrrrrrrrr}
\toprule
& \multicolumn{13}{c|}{Sub-Saharan Africa (SSA)} & \multicolumn{2}{c}{South-Asia} \\
\midrule
Language & ln & luo & nso & ny & om & sn & so & sw & umb & wo & xh & yo & zu & as & bn \\
\midrule 
Speech-to-Text Retrieval & 91.2 & 91.0 & 80.8 & 85.5 & 92.7 & 84.1 & 68.7 & 91.2 & 77.3 & 90.6 & 90.9 & 92.4 & 85.5 & 81.5 & 83.5 \\
Text-to-Speech Retrieval & 95.2 & 88.3 & 83.4 & 84.9 & 90.2 & 79.1 & 69.0 & 91.2 & 83.6 & 90.8 & 88.9 & 90.7 & 84.2 & 78.8 & 77.6 \\

\end{tabular}  

\begin{tabular}{l|rrrrrrrrrrrrrrr}
\toprule
& \multicolumn{12}{c|}{South-Asia (SA)} & \multicolumn{3}{c}{South-East Asia}\\
\midrule
Language & gu & hi & kn & ml & mr & ne & or & pa & sd & ta & te & ur & my & ceb & fil \\
\midrule 
Speech-to-Text Retrieval & 77.0 & 78.0 & 69.0 & 62.3 & 69.8 & 66.1 & 15.7 & 70.6 & 71.8 & 58.0 & 73.5 & 70.6 & 2.4 & 79.8 & 73.1 \\
Text-to-Speech Retrieval & 73.6 & 66.0 & 67.1 & 57.1 & 70.0 & 62.4 & 18.6 & 70.2 & 64.1 & 54.5 & 67.2 & 33.1 & 2.6 & 75.6 & 69.5 \\

\end{tabular}  

\begin{tabular}{l|rrrrrrrrrrrrrrr}
\toprule
& \multicolumn{8}{c|}{South-East Asia (SEA)} & \multicolumn{4}{c|}{CJK} & & & \\
\midrule
Language & id & jv & km & lo & ms & mi & th & vi & yue & cmn & ja & ko & All & & \\
\midrule 
Speech-to-Text Retrieval & 79.6 & 78.0 & 42.1 & 37.0 & 77.7 & 64.7 & 3.2 & 64.5 & 2.4 & 5.4 & 5.8 & 5.2 & 76.9 &  \\
Text-to-Speech Retrieval & 75.3 & 72.5 & 49.3 & 47.7 & 73.7 & 74.8 & 3.8 & 64.4 & 2.7 & 5.1 & 8.3 & 2.9 & 74.4 & \\
\bottomrule
\end{tabular}

\end{table*}
}

\newcommand{\insertretrievalseenvsunseen}{
 \begin{table}[bht!]
        \begin{center}
            \captionof{table}{\textbf{  Cross-modal Speech-Text Retrieval }- FLEURS massively multilingual Speech-to-Text and Text-to-Speech retrieval baselines, reporting \% P@1 ($\uparrow$) score, by geographical group.
            \label{tab:fleurs_retrieval}}
            \resizebox{\linewidth}{!}{
            \begin{tabular}[t]{l|ccccccc|c}
            \toprule
            {Task} & WE & EE & CMN & SSA & SA & SEA & CJK & Avg. \\
                \midrule
                \midrule
                \multicolumn{9}{l}{\it P@1 for all languages} \\
                \midrule
                \multicolumn{1}{l|}{\# languages} & 25 & 16 & 12 & 20 & 14 & 11 & 4 & 102 \\
                \midrule
                Speech-to-Text Retrieval & 87.6 & 91.1 & 79.4 & 83.9 & 67.7 & 54.8 & 4.7 & 76.9 \\
                Text-to-Speech Retrieval & 83.7 & 88.3 & 77.1 & 83.5 & 61.4 & 55.4 & 4.7 & 74.4 \\
                \bottomrule
            \end{tabular}
            }
        \end{center}
    \end{table}
}

\begin{document}

\maketitle
\begin{abstract}
We introduce FLEURS, the Few-shot Learning Evaluation of Universal Representations of Speech benchmark. FLEURS is an n-way parallel speech dataset in 102 languages built on top of the machine translation FLoRes-101 benchmark, with approximately 12 hours of speech supervision per language. FLEURS can be used for a variety of speech tasks, including Automatic Speech Recognition (ASR), Speech Language Identification (Speech LangID), Translation and Retrieval. In this paper, we provide baselines for the tasks based on multilingual pre-trained models like mSLAM. The goal of FLEURS is to enable speech technology in more languages and catalyze research in low-resource speech understanding.\footnote{The dataset is publicly available via TFDS at \url{https://tensorflow.org/datasets/catalog/xtreme_s} and Huggingface at \url{https://hf.co/datasets/google/xtreme_s}.\\ \textsuperscript{\dag} Equal Contributions. \textsuperscript{\ddag} Equal Advising Contributions. Work done while Alexis and Siddharth were at Google.} 

\end{abstract}

\section{Introduction}
Speech technology has seen rapid development in the past few years, from the use of self-attention models~\cite{vaswani2017transformer,gulati2020conformer} to pre-training approaches~\cite{baevski2020wav,conneau2021xlsr}. Methods like wav2vec 2.0 have demonstrated strong performance on LibriSpeech~\cite{panayotov2015librispeech}, in particular in the few-shot learning scenario with only 10 minutes of labeled data~\cite{xu2020selftraining}. The recent scaled-up multilingual wav2vec 2.0 model, dubbed XLS-R~\cite{babu2021xlsr}, has expanded similar few-shot capabilities to many more languages, including low-resource ones. By leveraging large-scale pretraining datasets like Multilingual LibriSpeech~\cite{pratap2020mls} and VoxPopuli~\cite{wang2021voxpopuli}, XLS-R also provides representations that can be used across tasks, with significant gains over previous baselines on speech recognition, translation and classification. More recently, mSLAM~\cite{bapna2022mslam}, a joint speech and text multilingual pretrained model, outperformed XLS-R on speech translation and ASR and improved over speech-only baselines on Speech-LangID.

This progress was made possible by the release of both large-scale pretraining and evaluation datasets like Multilingual LibriSpeech~\cite{pratap2020mls}, VoxPopuli~\cite{wang2021voxpopuli}, CoVoST-2~\cite{wang2020covost2}, CommonVoice~\cite{ardila2019common} and the re-use of existing datasets like BABEL~\cite{gales2014babel}. These datasets provide evaluations for speech recognition and speech translation in between 8 and 60 languages each. VoxLingua107 ~\cite{valk2020voxlingua107} also provides a Speech LangID classification dataset in 107 languages.
\insertcorpora

In machine translation, the release of new benchmarks like FLoRes-101~\cite{goyal2021flores} has enabled advances in publicly available massively multilingual machine translation systems~\cite{fan2021beyond}. With FLEURS, we hope to provide a resource that could catalyze research towards building massively multilingual speech and text representations and their evaluation on a variety of tasks. While there are a few other datasets containing n-way parallel speech and text, including Europarl-ST~\cite{iranzo2020europarl}, MuST-C~\cite{cattoni2021must}, mTEDx~\cite{salesky2021multilingual} and the CVSS corpus~\cite{jia2022cvss}, to the best of our knowledge, FLEURS is the only dataset spanning over 100 languages enabling research on a diverse set of languages. There are a few key properties of FLEURS that are important to note:
\begin{itemize}
    \item FLEURS contains n-way parallel speech and text in 102 languages, see Table~\ref{tab:langs} for a full list;
    \item FLEURS provides natural human speech and high quality transcripts for each language with strong quality control;
    \item FLEURS uses a bottom up approach of collecting spoken utterances for aligned segments, while most other datasets are aligned at a document level with automatic segmentation and alignment for segments;
\end{itemize}
FLEURS is well-suited for several downstream tasks including ASR, Speech-to-Text and Speech-to-Speech Translation, Speech LangID, and Multilingual Speech-to-Speech and Speech-to-Text Retrieval. We compare FLEURS to existing common public multilingual corpora in Table~\ref{tab:corpora}.

In addition to the describing the dataset, we provide baselines for Speech-LangID, ASR and Speech-Text retrieval (both Speech-to-Text and Text-to-Speech retrieval) by fine-tuning the mSLAM~\cite{bapna2022mslam} and multilingual w2v-BERT~\cite{chung2021w2vbert,bapna2022mslam} baselines on these tasks.

\section{Dataset}

\subsection{Speech Data Collection}
Our data collection process is simple. We start with the FLoRes-101 dataset.\footnote{Note: For clarity we have renamed FLoRes ``Chinese (Simp)'' to ``Mandarin Chinese'' (code ``cmn'') and ``Chinese (Trad)'' to ``Cantonese Chinese'' (code "yue").} FLoRes-101 contains 3001 sentences extracted from English Wikipedia and these sentences have been translated in 101 languages by human translators.
Because the test set of FLoRes-101 is not publicly available, we only use the dev and devtest sets, which contain 2009 sentences in total. We split those sentences into new train, dev and test splits, with 1509, 150 and 350 sentences respectively. For each sentence in the 102 languages (101 counted in FLoRes plus English, see Table~\ref{tab:langs}), we collected three recordings by three different native speakers, imposing a balance in terms of sex ratio of at least 30/70\%, when possible.


After this first recording step, each recording is evaluated by additional workers to assesses whether the recording corresponds to the input sentence.
Invalid recordings are discarded, leaving us between zero and three recordings per sentence in the final dataset. In the first version of the dataset, about 21.5\% of the sentences are missing because none of the three recordings were validated. We plan to fill these gaps in the future versions of the dataset. All recordings are kept as they-are, either from quiet or noisy environment, without any data augmentations ($e.g.$ SpecAugment~\cite{park2019specaugment}, speed perturbation~\cite{speedperturbation}, simulated reverberation~\cite{reverberation}, etc.). The speech sampling rate is 16kHz. 

We then split the collected data into train, development (dev) and test sets with disjoint speakers between train/dev and test, with a target ratio of 7:1:2. All the segments are within 30 seconds.

\subsection{Textual Data}
For source transcripts, we reuse the transcripts produced by human translators from \cite{goyal2021flores}. We maintain the English translated transcripts, which are useful for tasks such as multi-modal speech translation evaluations. 

Among the various possible modeling units ($e.g.$ character or sentence-pieces) for massively multilingual ASR, a universal vocabulary of characters requires the least resources to build, and better matches a common evaluation metric ($i.e.$ character level error rate). We adopt it as our modeling and evaluation unit.

The variety of orthographic symbols of languages complicates the tokenization process. For example, Chinese text in both traditional and simplified scripts does not have space between tokens. Depending on the transcribers, Japanese and Korean may or may not contain space irregularly. To ease the pain for other researchers and facilitate apple-to-apple comparisons and reproducibility, we provide the tokenized versions of the sentences. We apply NFC and then FST normalization to each sentence, lower-case, normalize and remove punctuations. We also split words into characters, and define the token $|$ as word boundaries.
For each sentence, three versions are provided: the original raw transcript (\textsc{src\_raw}), the preprocessed version (\textsc{src\_norm}) and its character-based version (\textsc{src\_char}), which should be used for ASR.

\subsection{Taxonomy and Statistics}
\begin{figure}[t]
  \centering
  \includegraphics[width=\linewidth]{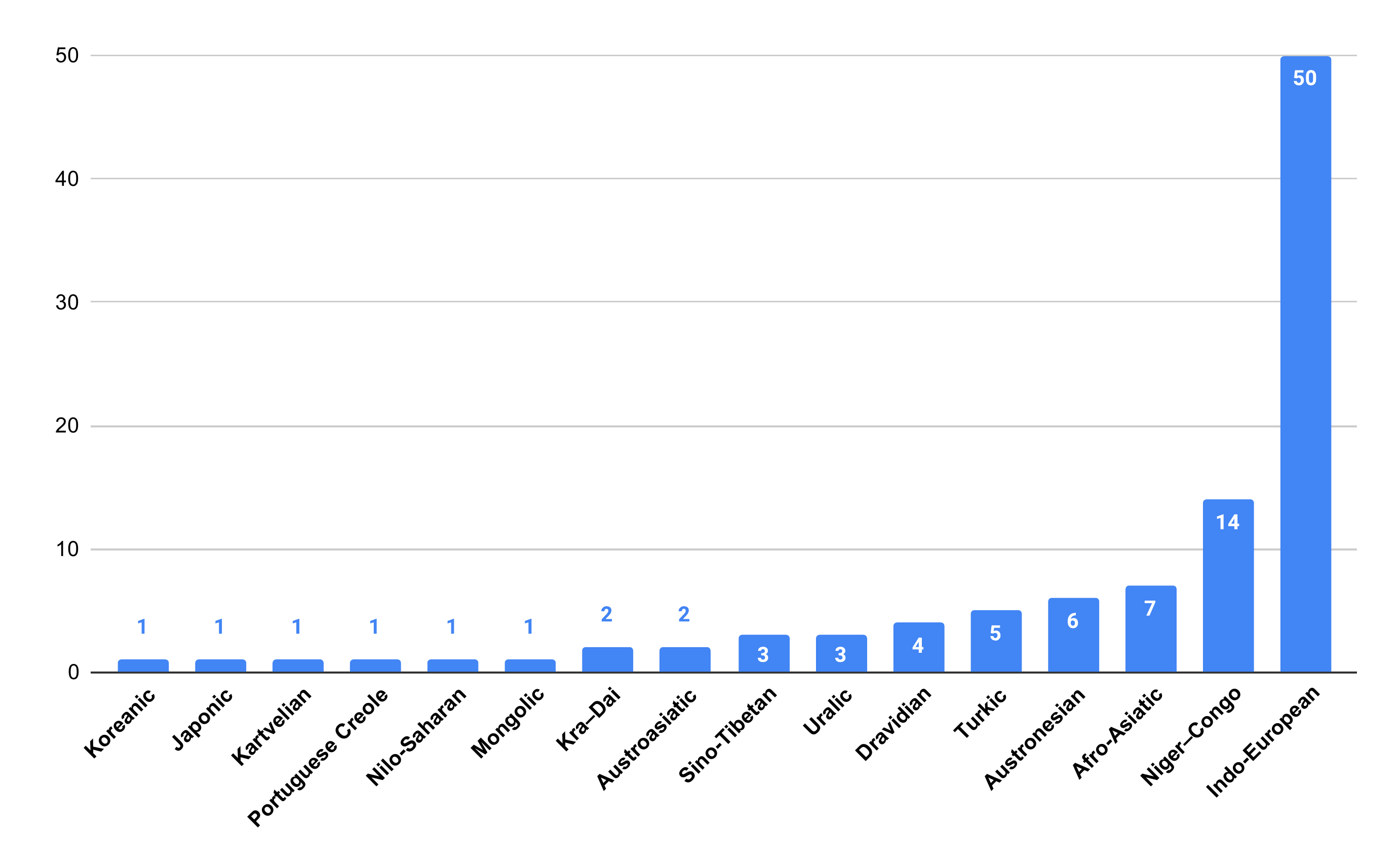}
  \caption{Distributions of language families in FLEURS (y-axis is the count).}
  \label{fig:languagefamilies}
\end{figure}

\begin{figure}[t]
  \centering
  \includegraphics[width=\linewidth]{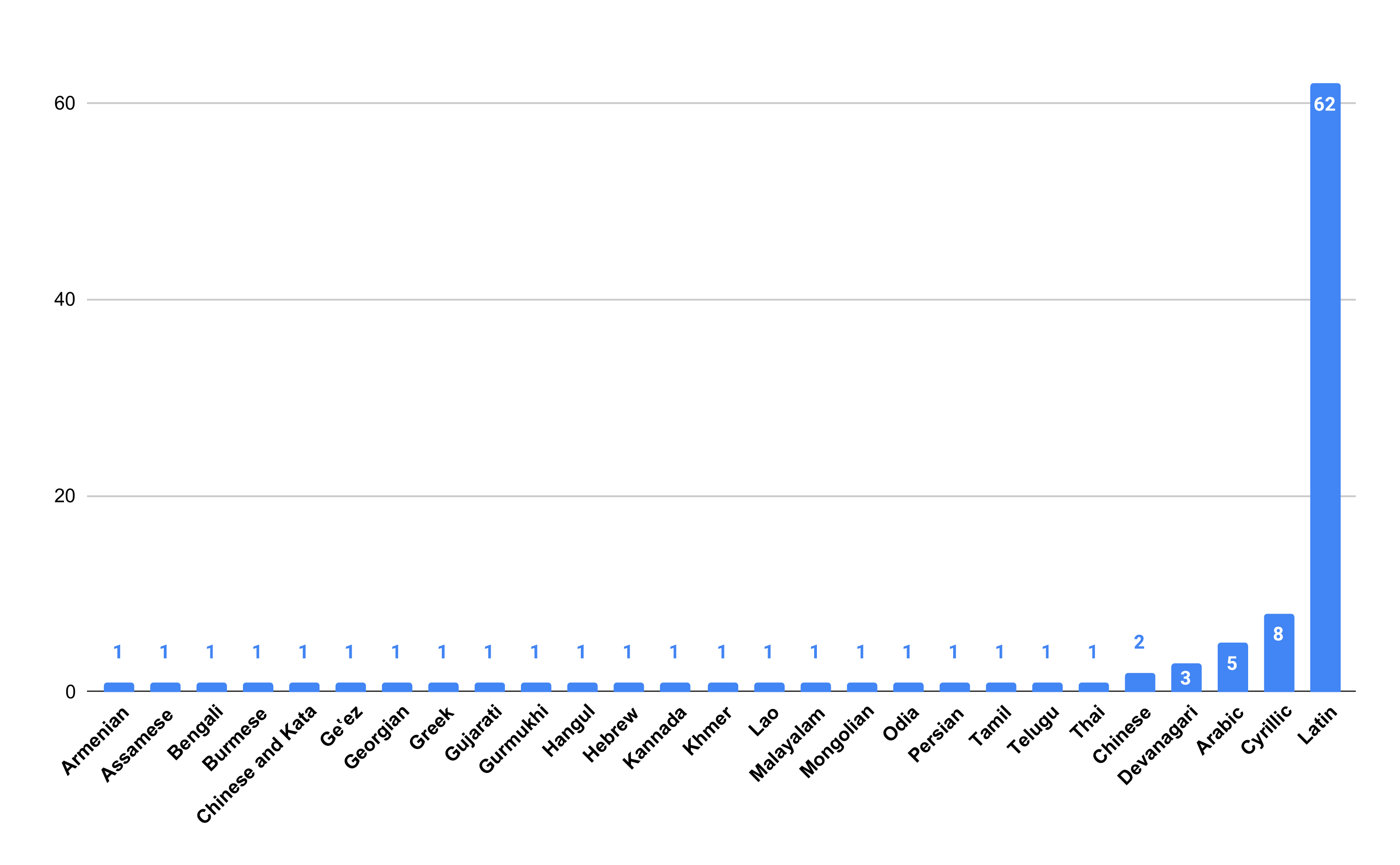}
  \caption{Distributions of writing systems in FLEURS (y-axis is the count).}
  \label{fig:writingsystems}
\end{figure}

There are multiple ways to categorize languages. 
The languages of FLEURS cover 17 language families 
(distribution shown in Figure~\ref{fig:languagefamilies}), and 27 unique writing systems (distribution shown in Figure~\ref{fig:writingsystems}). By construction, FLoRes sentences also cover a diversity in domains, including nature, politics, science, travel, sports etc. Each sentence also has an associated integer "index" between 1 and 2009, which can be used to recover the n-way parallelism from one language to another (i.e. sentence $i$ in language $A$ is the translation of sentence $i$ in language $B$). 
We grouped languages along with their geographical areas: Western European (WE), Eastern  Europe (EE), Central-Asia/Middle-East/North-Africa (CMN), Sub-Saharan Africa (SSA), South Asia (SA), South-East Asia (SEA) and Chinese, Japanese, and Korean (CJK) languages.  In Table~\ref{tab:fleursbasics}, we present the basic statistics of FLEURS data per geographical groups; see Table~\ref{tab:langs} for a full list of languages.






\insertfleursbasics

\section{Task Baselines}
\subsection{Experimental Setup}
FLEURS enables evaluations for several core speech tasks. In this paper, we focus on speech recognition, speech language identification and speech-text retrieval.
{\color{black} Training a giant model from scratch on the FLEURS dataset will easily overfit. Fine-tuning on limited supervised data from a model pre-trained on a vast amount of (unsupervised) data has achieved state-of-the-art performances in recent studies~\cite{zhang2021bigssl, chung2021w2vbert}. Consequently, we adopt the pre-training and fine-tuning methodology to build our massively multilingual baselines.} \\
\textbf{Multilingual Pre-trained} Multilingual pre-trained models have achieved significant gains in a range of NLP and ASR tasks. We initialize fine-tuning from a 600M parameter wav2vec-BERT~\cite{chung2021w2vbert} model, which has been pre-trained on 429k hours unlabeled speech data in 51 languages pooling from VoxPopuli~\cite{wang2021voxpopuli}, MLS~\cite{pratap2020mls}, CommonVoice~\cite{ardila2019common} and BABEL~\cite{gales2014babel}. Pre-training for this baseline (dubbed \wvbert{}) is \emph{speech-only}.
\\
\textbf{Multilingual Multimodal Pre-trained} In addition to pre-training on multilingual speech, pre-training on speech and text data simultaneously allows for transfer learning across the two modalities. We explore fine-tuning from a multilingual model which has been pre-trained with the same speech data, and with more than 10TiB of unlabeled text from 101 languages~\cite{xue2020mt5}. The pre-trained model capacity is also of 600M parameters (dubbed as \mctcb{0.6}).

More information about the baselines, including fine-tuning details can be found in \cite{bapna2022mslam}.

\subsection{\color{black} Seen Languages and Unseen Languages}
The languages for which speech data was presented to pre-training are referred to as \emph{seen} languages. There are 54 seen languages:

\begin{itemize}
    \item \textbf{WE}: Catalan (ca), Croatian (hr), Danish (da), Dutch (nl), American English (en), Finnish (fi), French (fr), German (de), Greek (el), Hungarian (hu), Irish (ga), Italian (it), Latin American Spanish (es), Maltese (mt), Portuguese (pt), Swedish (sv), Welsh (cy)
    \item \textbf{EE}: Bulgarian (bg), Czech (cs), Estonian (et), Georgian (ka), Latvian (lv), Lithuanian (lt), Polish (pl), Romanian (ro), Russian (ru), Slovak (sk), Slovenian (sl), Ukrainian (uk)
    \item \textbf{CMN}: Arabic (ar), Kazakh (kk), Kyrgyz (ky), Mongolian (mn), Pashto (ps), Persian (fa), Tajik (tg), Turkish (tr)
    \item \textbf{SSA}:  Ganda (lg), Swahili (sw), Zulu (zu)
    \item \textbf{SA}: Assamese (as), Bengali (bn), Hindi (hi), Oriya (or), Punjabi (pa), Tamil (ta), Telugu (te)
    \item \textbf{SEA}: Cebuano (ceb), Indonesian (id), Lao (lo), Thai (th), Vietnamese (vi)
    \item \textbf{CJK}: Cantonese (yue), Japanese (ja), Mandarin (cmn)
\end{itemize}

Languages which do not have any speech data presented to pre-training are referred to as \emph{unseen} languages. There are 48 unseen languages:
\begin{itemize}
    \item \textbf{WE}: Asturian (ast), Bosnian (bs), Galician (gl), Icelandic (is), Kabuverdianu (kea), Luxembourgish (lb), Norwegian (nb), Occitan (oc)
    \item \textbf{EE}: Armenian (hy), Belarusian (be), Macedonian (mk), Serbian (sr)
    \item \textbf{CMN}: Azerbaijani (az), Hebrew (he), Sorani-Kurdish (ckb), Uzbek (uz)
    \item \textbf{SSA}: Afrikaans(af), Amharic (am), Fula (ff), Hausa (ha), Igbo (ig), Kamba (kam), Lingala (ln), Luo (luo), Northern-Sotho (nso), Nyanja (ny), Oromo (om), Shona (sn), Somali (so), Umbundu (umb), Wolof (wo), Xhosa (xh), Yoruba (yo)
    \item \textbf{SA}: Gujarati (gu), Kannada (kn), Malayalam (ml), Marathi (mr), Nepali (ne), Sindhi (sd), Urdu (ur)
    \item \textbf{SEA}: Filipino (fil), Javanese (jv), Khmer (km), Malay (ms), Maori (mi), Burmese (my)
    \item \textbf{CJK}: Korean (ko)
\end{itemize}

\noindent Text in pre-training includes most of the above, except: as, ast, bg, bs, ff, he, hr, kam, kea, lb, lg, ln, lo, luo, nb, nso, oc, om, or, umb and wo. 
\section{Downstream Tasks}
\subsection{Speech Recognition}
%
We add two LSTM\cite{HochSchm97} layer to  fine-tune our pre-trained models for ASR, using a CTC~\cite{graves2006connectionist} loss. {The baselines use a 6100-character vocabulary built from \textsc{src\_norm}. We do not include meta information of language identification labels in modeling, and there is no language model used for hypothesis scoring.  We fine-tune on all 102 locales, and report results for multilingual fine-tuning from both speech-only pre-trained and speech-text pre-trained models. Our finetuning parameters follow~\cite{bapna2022mslam}. We evaluate fine-tuned ASR models for all locales in terms of \% character error rate (CER). Experiments show that fine-tuning from multimodal pre-training is slightly worse than fine-tuning from speech-only pre-training (similar to the patterns observed in~\cite{bapna2022mslam}), particularly lagging behind for the \emph{unseen} languages.} Most gains of multimodal pre-training are observed in WE, EE and CMN groups, suggesting that multimodal pre-training can be promising for certain languages, especially those where relatively large amounts of unlabeled speech, text and paired data were available during pre-training.

\subsubsection{Correlation with Language Groups}
\label{sec:asr_corr}
We observe in Table~\ref{tab:fleurs} that results are much better on the Western European group (with 10.7 and 10.6 average CER) than on other groups like Sub-Saharan African (15.6 and 16.4 average CER) or South Asian (17.4 and 19.2), which can be explained in part due to the larger amounts of unlabeled data in WE languages from MLS and VoxPopuli. Reducing the gaps across geographical groups is an important research direction for future work.

{\color{black} As presented in Table~\ref{tab:fleurs}, fine-tuning from a speech-text pre-trained model leads to 0.5\% regression in CER as compared to fine-tuning from a speech-only pre-trained model. We observed most degradation in SA, SSA and CJK, three geographical groups known for mixing textual systems.
For the model initialized from speech-only pre-training, EE and WE are the groups with the lowest average CERs, CMN, SSA, SA and SEA observe moderate CERs, while CJK gets the highest group CER. Further breaking down the distributions of error types, substitution errors are dominating across all the groups, which is a common error pattern in multilingual ASR~\cite{li2020universal}, especially when no explicit language id information was incorporated. CJK languages are known for the vast number of homophones in speech, which adds difficulties in selecting the correct character without aid from language models. For SA, the group seeing the second highest substitution error rate, most substitutions come from Urdu. This is due to being essentially acoustically similar to Hindi, many Urdu utterances have been transliterated into Devanagari script. In other geographical groups, mis-recognized characters from another writing system are observed less frequently. A potential solution is to include the language specific information and utilize language model fusion.}


\insertfleursasrseenvsunseen

\subsubsection{Generalizability}

\textbf{\color{black} Languages Seen by Pre-training}
{\color{black} A majority of the languages which have been seen in pre-training observed a test CER lower than average. While most of the languages with high CERs suffer from relatively high substitution error rates, ga and lo are also undermined by high insertion and deletion error rates.}
\\
\\
\textbf{\color{black} Languages Unseen by Pre-training}
{\color{black} For languages which were not seen in pre-training, a large fraction observed a test CER worse than global average due to fine-tuning on very limited amount of supervised data. These observations align with previous findings in \cite{babu2021xlsr}.}

For the unseen languages which achieved a lower CER than average, 15 use Latin script based writing systems, while 3 use Cyrillic. Languages with Latin-scripts have a higher average CER than those using Cyrillic. Interestingly, unseen languages which use scripts other than Latin or Cyrillic can obtain good CER: ml\_in (Malayalam), kn\_in (Kannada),  gu\_in (Gujarati), ne\_np (Devanagari), my\_mm (Burmese), ckb\_iq (Arabic). The success in recognizing unseen Malayalam, Kannada, Gujarati, Nepali can potentially be attributed to the Indian languages seen during pre-training (bn\_in, te\_in, pa\_in, as\_in, ta\_in) from pre-training. 



\subsection{Speech Language Identification}
We fine-tune our models on Speech LangID classification following \cite{bapna2022mslam}. As shown in Table~\ref{tab:fleurssid}, fine-tuning from mSLAM obtains 73.3\% macro-average accuracy on FLEURS LangID, while fine-tuning from \wvbert{} obtains 71.4\% respectively. The group average accuracy decreases in the order of: $\text{CJK} > \text{WE} > \text{EE} > \text{CMN} > \text{SEA} > \text{SSA} > \text{SA}$. This could be due to the following reasons: (1) there are only four languages in the CJK group, which are relatively easy to distinguish from each other and from languages in the other groups; (2) Most of the data seen during pre-training is from Western European and Eastern European languages; (3) CMN, SEA, SSA and SA are geographical regions which are known for language diversity, but with limited amounts of publicly available pre-training data.

We note that on FLEURS-LangID, speakers are different among the train and dev/test sets. Avoiding over-fitting on speaker ID for the LangID task is essential for obtaining good performance.

\insertsc

\subsection{Cross-modal Speech-Text Retrieval}
As user interactions with machines moves beyond typing textual queries, multi-modal retrieval of documents across the web is of growing interest~\cite{witbrock1997speech, ishikawa2004speech, lee2015spoken}. The ability to build fixed-sized vector representations for queries be it speech, text or images, allows for building such content retrieval models~\cite{jain2021mural,duquenne2021multimodal}. FLEURS is a rich multi-lingual, multi-modal, n-way parallel dataset which we hope will act as a benchmark to accelerate research in this field. FLEURS can act as a test-bench for various kinds of speech-text retrieval scenarios like speech-to-speech, text-to-text, speech-to-text, and text-to-speech retrieval along with testing cross-lingual and zero-shot capabilities. 

As part of our multilingual baseline, we study the efficacy of pre-trained models towards learning fixed sized representations for both speech-to-text and text-to-speech retrieval. 
Given the multi-modal nature of the task, we only fine-tune our multi-lingual multi-modal pretrained model, i.e., mSLAM (0.6B), on the training set from all languages in FLEURS. Following ~\cite{yang2019improving,feng2020language}, cross-modal embeddings are
trained using the additive margin softmax loss with in-batch negative sampling. We add bi-directional loss for retrieving speech given a text query and vice-versa \cite{yang2019improving}. We obtain embeddings for the normalized text (\textsc{src\_norm}) for all languages.

For evaluating the speech-to-text retrieval task, we report the \% Precision at 1 (P@1) retrieval score of retrieving the correct text segment given a speech query from a database of in-domain textual keys collected from the FLEURS test set. Similarly, for text-to-speech retrieval task, we report the P@1 retrieval score of retrieving any of the speakers who speaks the correct textual query. We report results for both retrieving text segments from speech queries and speech segments from textual queries. The summary tables for each geographical group can be found in Table \ref{tab:fleurs_retrieval} for both speech-to-text and text-to-speech retrieval. The detailed retrieval score for each language can be found in Table \ref{tab:fleurs_full_retrieval}. 

We observe an average P@1 of 76.9\% for speech-to-text retrieval and a P@1 of 74.4\% for text-to-speech retrieval. We observe that P@1 for seen languages in almost all geographical groups (except SA and SEA) is higher than their unseen counterparts, as is the case with speech recognition and language identification. In particular, we notice a steep degradation in the retrieval performance on CJK languages. We anticipate this to be the result of tokenization mismatch between the fine-tuning and the pre-training regime. 

From Table \ref{tab:fleurs_full_retrieval}, we can observe some interesting language specific peculiarities in the retrieval performance. For example, while Odia (or) is seen in speech, it is unseen in text since it is not present in the mc4 corpus \cite{xue2020mt5} on which the mSLAM model was trained. This is exacerbated by Odia's unique script ~\cite{mohapatra2015ohcs}, which leads to the language being unrepresented in the tokenizer. On the other hand, Urdu (ur) is seen in the text pre-training but unseen in speech. This is interesting because Urdu performs considerably worse in Text-to-Speech retrieval compared to Speech-to-Text. We believe this is because Urdu is phonetically close to other SA languages like Hindi, consistent with our observations in Section \ref{sec:asr_corr}, making it hard for the model to disambiguate speech without pre-training data in the speech modality. 

\insertretrievalseenvsunseen



\section{Conclusion}
In this paper, we introduced FLEURS a new dataset for Few-shot Learning Evaluation of Universal Representations of Speech, in 102 languages. FLEURS is an n-way parallel speech dataset that can be used to evaluate speech recognition, translation, classification and retrieval methods. By building up baseline ASR, language identification and retrieval systems on it, we show that it is especially suited to evaluate data-efficient multilingual pre-trained representations of speech (and text). We hope this dataset will catalyze research in few-shot understanding in many languages, enabling progress towards building speech technologies for everyone.

\bibliographystyle{IEEEtran}
\bibliography{template.bbl}

\begin{thebibliography}{10}
\providecommand{\url}[1]{#1}
\csname url@samestyle\endcsname
\providecommand{\newblock}{\relax}
\providecommand{\bibinfo}[2]{#2}
\providecommand{\BIBentrySTDinterwordspacing}{\spaceskip=0pt\relax}
\providecommand{\BIBentryALTinterwordstretchfactor}{4}
\providecommand{\BIBentryALTinterwordspacing}{\spaceskip=\fontdimen2\font plus
\BIBentryALTinterwordstretchfactor\fontdimen3\font minus
  \fontdimen4\font\relax}
\providecommand{\BIBforeignlanguage}[2]{{%
\expandafter\ifx\csname l@#1\endcsname\relax
\typeout{** WARNING: IEEEtran.bst: No hyphenation pattern has been}%
\typeout{** loaded for the language `#1'. Using the pattern for}%
\typeout{** the default language instead.}%
\else
\language=\csname l@#1\endcsname
\fi
#2}}
\providecommand{\BIBdecl}{\relax}
\BIBdecl

\bibitem{vaswani2017transformer}
A.~Vaswani, N.~Shazeer, N.~Parmar, J.~Uszkoreit, L.~Jones, A.~N. Gomez,
  L.~Kaiser, and I.~Polosukhin, ``Attention is all you need,'' in \emph{Proc.
  of NIPS}, 2017.

\bibitem{gulati2020conformer}
A.~Gulati, J.~Qin, C.-C. Chiu, N.~Parmar, Y.~Zhang, J.~Yu, W.~Han, S.~Wang,
  Z.~Zhang, Y.~Wu, and R.~Pang, ``Conformer: Convolution-augmented transformer
  for speech recognition,'' \emph{Proc. of Interspeech}, 2020.

\bibitem{baevski2020wav}
A.~Baevski, Y.~Zhou, A.~Mohamed, and M.~Auli, ``wav2vec 2.0: {A} framework for
  self-supervised learning of speech representations,'' in \emph{Proc. of
  NeurIPS}, 2020.

\bibitem{conneau2021xlsr}
A.~Conneau, A.~Baevski, R.~Collobert, A.~Mohamed, and M.~Auli, ``Unsupervised
  cross-lingual representation learning for speech recognition,'' in
  \emph{Proc. of Interspeech}, 2021.

\bibitem{panayotov2015librispeech}
V.~Panayotov, G.~Chen, D.~Povey, and S.~Khudanpur, ``Librispeech: an asr corpus
  based on public domain audio books,'' in \emph{Proc. of ICASSP}.\hskip 1em
  plus 0.5em minus 0.4em\relax IEEE, 2015, pp. 5206--5210.

\bibitem{xu2020selftraining}
Q.~Xu, A.~Baevski, T.~Likhomanenko, P.~Tomasello, A.~Conneau, R.~Collobert,
  G.~Synnaeve, and M.~Auli, ``Self-training and pre-training are complementary
  for speech recognition,'' in \emph{Proc. of ICASSP}, 2020.

\bibitem{babu2021xlsr}
A.~Babu, C.~Wang, A.~Tjandra, K.~Lakhotia, Q.~Xu, N.~Goyal, K.~Singh, P.~von
  Platen, Y.~Saraf, J.~Pino, A.~Baevski, A.~Conneau, and M.~Auli, ``{XLS-R}:
  Self-supervised cross-lingual speech representation learning at scale,''
  \emph{arXiv preprint arXiv:2111.09296}, 2021.

\bibitem{pratap2020mls}
V.~Pratap, Q.~Xu, A.~Sriram, G.~Synnaeve, and R.~Collobert, ``{MLS}: A
  large-scale multilingual dataset for speech research,'' in \emph{Proc. of
  Interspeech}, 2020.

\bibitem{wang2021voxpopuli}
C.~Wang, M.~Riviere, A.~Lee, A.~Wu, C.~Talnikar, D.~Haziza, M.~Williamson,
  J.~Pino, and E.~Dupoux, ``{V}ox{P}opuli: A large-scale multilingual speech
  corpus for representation learning, semi-supervised learning and
  interpretation,'' in \emph{Proc. of ACL}, 2021.

\bibitem{bapna2022mslam}
\BIBentryALTinterwordspacing
A.~Bapna, C.~Cherry, Y.~Zhang, Y.~Jia, M.~Johnson, Y.~Cheng, S.~Khanuja,
  J.~Riesa, and A.~Conneau, ``{mSLAM}: Massively multilingual joint
  pre-training for speech and text,'' \emph{CoRR}, vol. abs/2202.01374, 2022.
  [Online]. Available: \url{https://arxiv.org/abs/2202.01374}
\BIBentrySTDinterwordspacing

\bibitem{wang2020covost2}
C.~Wang, A.~Wu, and J.~Pino, ``{CoVoST} 2 and massively multilingual
  speech-to-text translation,'' \emph{arXiv}, 2020.

\bibitem{ardila2019common}
R.~Ardila, M.~Branson, K.~Davis, M.~Henretty, M.~Kohler, J.~Meyer, R.~Morais,
  L.~Saunders, F.~M. Tyers, and G.~Weber, ``{Common Voice}: A
  massively-multilingual speech corpus,'' \emph{Proc. of LREC}, 2020.

\bibitem{gales2014babel}
M.~J.~F. Gales, K.~M. Knill, A.~Ragni, and S.~P. Rath, ``Speech recognition and
  keyword spotting for low-resource languages: Babel project research at
  cued,'' in \emph{n Spoken Language Technologies for Under-Resourced
  Languages}, 2014.

\bibitem{valk2020voxlingua107}
J.~Valk and T.~Alumäe, ``{VoxLingua107}: a dataset for spoken language
  recognition,'' in \emph{Proc. of SLT}, 2020.

\bibitem{black2019cmu}
A.~W. Black, ``{CMU} wilderness multilingual speech dataset,'' in \emph{ICASSP
  2019-2019 IEEE International Conference on Acoustics, Speech and Signal
  Processing (ICASSP)}.\hskip 1em plus 0.5em minus 0.4em\relax IEEE, 2019, pp.
  5971--5975.

\bibitem{iranzo2020europarl}
J.~Iranzo-S{\'a}nchez, J.~A. Silvestre-Cerda, J.~Jorge, N.~Rosell{\'o},
  A.~Gim{\'e}nez, A.~Sanchis, J.~Civera, and A.~Juan, ``{Europarl-ST}: A
  multilingual corpus for speech translation of parliamentary debates,'' in
  \emph{ICASSP 2020-2020 IEEE International Conference on Acoustics, Speech and
  Signal Processing (ICASSP)}.\hskip 1em plus 0.5em minus 0.4em\relax IEEE,
  2020, pp. 8229--8233.

\bibitem{cattoni2021must}
R.~Cattoni, M.~A. Di~Gangi, L.~Bentivogli, M.~Negri, and M.~Turchi, ``Must-c: A
  multilingual corpus for end-to-end speech translation,'' \emph{Computer
  Speech \& Language}, vol.~66, p. 101155, 2021.

\bibitem{salesky2021multilingual}
E.~Salesky, M.~Wiesner, J.~Bremerman, R.~Cattoni, M.~Negri, M.~Turchi, D.~W.
  Oard, and M.~Post, ``The multilingual {TEDx} corpus for speech recognition
  and translation,'' \emph{arXiv preprint arXiv:2102.01757}, 2021.

\bibitem{jia2022cvss}
Y.~Jia, M.~T. Ramanovich, Q.~Wang, and H.~Zen, ``{CVSS} corpus and massively
  multilingual speech-to-speech translation,'' \emph{arXiv preprint
  arXiv:2201.03713}, 2022.

\bibitem{goyal2021flores}
N.~Goyal, C.~Gao, V.~Chaudhary, P.-J. Chen, G.~Wenzek, D.~Ju, S.~Krishnan,
  M.~Ranzato, F.~Guzman, and A.~Fan, ``The {FLORES-101} evaluation benchmark
  for low-resource and multilingual machine translation,'' \emph{arXiv preprint
  arXiv:2106.03193}, 2021.

\bibitem{fan2021beyond}
A.~Fan, S.~Bhosale, H.~Schwenk, Z.~Ma, A.~El-Kishky, S.~Goyal, M.~Baines,
  O.~Celebi, G.~Wenzek, V.~Chaudhary \emph{et~al.}, ``Beyond english-centric
  multilingual machine translation,'' \emph{Journal of Machine Learning
  Research}, vol.~22, no. 107, pp. 1--48, 2021.

\bibitem{chung2021w2vbert}
Y.-A. Chung, Y.~Zhang, W.~Han, C.-C. Chiu, J.~Qin, R.~Pang, and Y.~Wu,
  ``{W2v-BERT}: Combining contrastive learning and masked language modeling for
  self-supervised speech pre-training,'' 2021.

\bibitem{park2019specaugment}
D.~S. Park, W.~Chan, Y.~Zhang, C.-C. Chiu, B.~Zoph, E.~D. Cubuk, and Q.~V. Le,
  ``{SpecAugment}: A simple data augmentation method for automatic speech
  recognition,'' \emph{Proc. Interspeech 2019}, pp. 2613--2617, 2019.

\bibitem{speedperturbation}
\BIBentryALTinterwordspacing
T.~Ko, V.~Peddinti, D.~Povey, and S.~Khudanpur, ``Audio augmentation for speech
  recognition.'' in \emph{INTERSPEECH}.\hskip 1em plus 0.5em minus 0.4em\relax
  ISCA, 2015, pp. 3586--3589. [Online]. Available:
  \url{http://dblp.uni-trier.de/db/conf/interspeech/interspeech2015.html#KoPPK15}
\BIBentrySTDinterwordspacing

\bibitem{reverberation}
T.~Ko, V.~Peddinti, D.~Povey, M.~L. Seltzer, and S.~Khudanpur, ``A study on
  data augmentation of reverberant speech for robust speech recognition,'' in
  \emph{2017 IEEE International Conference on Acoustics, Speech and Signal
  Processing (ICASSP)}.\hskip 1em plus 0.5em minus 0.4em\relax IEEE, 2017, pp.
  5220--5224.

\bibitem{zhang2021bigssl}
Y.~Zhang, D.~S. Park, W.~Han, J.~Qin, A.~Gulati, J.~Shor, A.~Jansen, Y.~Xu,
  Y.~Huang, S.~Wang, Z.~Zhou, B.~Li, M.~Ma, W.~Chan, J.~Yu, Y.~Wang, L.~Cao,
  K.~C. Sim, B.~Ramabhadran, T.~N. Sainath, F.~Beaufays, Z.~Chen, Q.~V. Le,
  C.-C. Chiu, R.~Pang, and Y.~Wu, ``{BigSSL}: Exploring the frontier of
  large-scale semi-supervised learning for automatic speech recognition,''
  \emph{arXiv}, 2021.

\bibitem{xue2020mt5}
L.~Xue, N.~Constant, A.~Roberts, M.~Kale, R.~Al-Rfou, A.~Siddhant, A.~Barua,
  and C.~Raffel, ``{mT5}: A massively multilingual pre-trained text-to-text
  transformer,'' \emph{arXiv preprint arXiv:2010.11934}, 2020.

\bibitem{HochSchm97}
S.~Hochreiter and J.~Schmidhuber, ``Long short-term memory,'' \emph{Neural
  Computation}, vol.~9, no.~8, pp. 1735--1780, 1997.

\bibitem{graves2006connectionist}
A.~Graves, S.~Fern{\'a}ndez, F.~Gomez, and J.~Schmidhuber, ``Connectionist
  temporal classification: labelling unsegmented sequence data with recurrent
  neural networks,'' in \emph{Proceedings of the 23rd international conference
  on Machine learning}, 2006, pp. 369--376.

\bibitem{li2020universal}
X.~Li, S.~Dalmia, J.~Li, M.~Lee, P.~Littell, J.~Yao, A.~Anastasopoulos, D.~R.
  Mortensen, G.~Neubig, A.~W. Black \emph{et~al.}, ``Universal phone
  recognition with a multilingual allophone system,'' in \emph{ICASSP 2020-2020
  IEEE International Conference on Acoustics, Speech and Signal Processing
  (ICASSP)}.\hskip 1em plus 0.5em minus 0.4em\relax IEEE, 2020, pp. 8249--8253.

\bibitem{witbrock1997speech}
M.~Witbrock and A.~G. Hauptmann, ``Speech recognition and information
  retrieval: Experiments in retrieving spoken documents,'' in \emph{Proceedings
  of the DARPA speech recognition workshop}, vol.~97, 1997.

\bibitem{ishikawa2004speech}
S.-y. Ishikawa, T.~Ikeda, K.~Miki, F.~Adachi, R.~Isotani, K.-I. Iso, and
  A.~Okumura, ``Speech-activated text retrieval system for multimodal cellular
  phones,'' in \emph{2004 IEEE International Conference on Acoustics, Speech,
  and Signal Processing}, vol.~1.\hskip 1em plus 0.5em minus 0.4em\relax IEEE,
  2004, pp. I--453.

\bibitem{lee2015spoken}
L.-s. Lee, J.~Glass, H.-y. Lee, and C.-a. Chan, ``Spoken content
  retrieval—beyond cascading speech recognition with text retrieval,''
  \emph{IEEE/ACM Transactions on Audio, Speech, and Language Processing},
  vol.~23, no.~9, pp. 1389--1420, 2015.

\bibitem{jain2021mural}
\BIBentryALTinterwordspacing
A.~Jain, M.~Guo, K.~Srinivasan, T.~Chen, S.~Kudugunta, C.~Jia, Y.~Yang, and
  J.~Baldridge, ``{MURAL}: Multimodal, multitask representations across
  languages,'' in \emph{Findings of the Association for Computational
  Linguistics: EMNLP 2021}.\hskip 1em plus 0.5em minus 0.4em\relax Punta Cana,
  Dominican Republic: Association for Computational Linguistics, Nov. 2021, pp.
  3449--3463. [Online]. Available:
  \url{https://aclanthology.org/2021.findings-emnlp.293}
\BIBentrySTDinterwordspacing

\bibitem{duquenne2021multimodal}
P.-A. Duquenne, H.~Gong, and H.~Schwenk, ``Multimodal and multilingual
  embeddings for large-scale speech mining,'' \emph{Advances in Neural
  Information Processing Systems}, vol.~34, 2021.

\bibitem{yang2019improving}
Y.~Yang, G.~H. {\'A}brego, S.~Yuan, M.~Guo, Q.~Shen, D.~Cer, Y.-H. Sung,
  B.~Strope, and R.~Kurzweil, ``Improving multilingual sentence embedding using
  bi-directional dual encoder with additive margin softmax,'' in \emph{IJCAI},
  2019.

\bibitem{feng2020language}
\BIBentryALTinterwordspacing
F.~Feng, Y.~Yang, D.~Cer, N.~Arivazhagan, and W.~Wang, ``Language-agnostic
  {BERT} sentence embedding,'' in \emph{Proceedings of ACL}.\hskip 1em plus
  0.5em minus 0.4em\relax Dublin, Ireland: Association for Computational
  Linguistics, May 2022, pp. 878--891. [Online]. Available:
  \url{https://aclanthology.org/2022.acl-long.62}
\BIBentrySTDinterwordspacing

\bibitem{mohapatra2015ohcs}
R.~K. Mohapatra, T.~K. Mishra, S.~Panda, and B.~Majhi, ``Ohcs: A database for
  handwritten atomic odia character recognition,'' in \emph{2015 Fifth National
  Conference on Computer Vision, Pattern Recognition, Image Processing and
  Graphics (NCVPRIPG)}.\hskip 1em plus 0.5em minus 0.4em\relax IEEE, 2015, pp.
  1--4.

\bibitem{Esch2022}
D.~van Esch, T.~Lucassen, S.~Ruder, I.~Caswell, and C.~E. Rivera, ``Writing
  system and speaker metadata for 2,800+ language varieties,'' in
  \emph{Proceedings of LREC}, 2022.

\end{thebibliography}
\newpage
\insertfleursfullasr
\insertfleursspeechtextretrieval

\appendix

\insertlanguages 
\end{document}